\documentclass[preprint,12pt]{elsarticle}
\usepackage{hyperref}
\usepackage{url}
\usepackage{natbib}
\usepackage{graphicx}
\usepackage{amsmath,amssymb,latexsym}
\usepackage{multirow}
\usepackage[applemac]{inputenc}

\usepackage{xspace}
\usepackage{tabularx}
\usepackage{multicol}
\usepackage{url}
\usepackage{wrapfig}
\usepackage[normalem]{ulem}
\usepackage{xcolor}

\journal{Pattern Recognition Letters}

\begin{document}

\begin{frontmatter}
\title{Alpha-Integration Pooling for Convolutional Neural Networks} 

\author{Hayoung Eom}
\author{Heeyoul Choi \fnref{fn1}}
\fntext[fn1]{Corresponding Author: heeyoul@gmail.com}
\address{Dept. of Information and Communication Engineering, \\
Handong Global University, \\ Pohang, South Korea 37554}

\begin{abstract}
Convolutional neural networks (CNNs) have achieved remarkable performance in many applications, especially in image recognition tasks. As a crucial component of CNNs, sub-sampling plays an important role for efficient training or invariance property, and max-pooling and arithmetic average-pooling are commonly used sub-sampling methods. In addition to the two pooling methods, however, there could be many other pooling types, such as geometric average, harmonic average, and so on. Since it is not easy for algorithms to find the best pooling method, usually the pooling types are assumed a priority, which might not be optimal for different tasks. In line with the deep learning philosophy, the type of pooling can be driven by data for a given task. In this paper, we propose {\it $\alpha$-integration pooling} ($\alpha$I-pooling), which has a trainable parameter $\alpha$ to find the type of pooling. $\alpha$I-pooling is a general pooling method including max-pooling and arithmetic average-pooling as a special case, depending on the parameter $\alpha$. Experiments show that $\alpha$I-pooling outperforms other pooling methods including max-pooling, in image recognition tasks. Also, it turns out that each layer has different optimal pooling type. 
 
\end{abstract}


\end{frontmatter}

\section{Introduction}
Deep learning has achieved remarkable performance in many applications, especially in image related tasks \cite{vgg,alexnet,resnext,inception}. In image recognition, convolutional neural networks (CNNs) are heavily used \cite{book_deeplearning}, which are based on a few components like convolutional layers and pooling (or sub-sampling) layers. The convolutional and pooling layers are motivated by neuroscientific discoveries \cite{hubel1962receptive}, where Hubel and Wiesel found that there are simple cells and complex cells in primary visual cortex. 

Simple cells respond best to bars with the particular orientations in the receptive field. Complex cells respond to appropriate orientations regardless of position or phase. In CNNs, these cells are modeled by convolutional and pooling layers, respectively. That is, convolutional layers catch local feature like simple cell, and pooling layers summarize the output of convolutional layers.
Pooling layers not only reduce the size of the feature-map but also extract features that are more robust against position or movement for object recognition like complex cell.

In most CNN models, max-pooling and arithmetic average-pooling are often selected for the pooling layers without further consideration. Max-pooling selects the highest value in the pooling window, and arithmetic average-pooling takes the arithmetic average in the window area. However, the two pooling methods may not be optimal. The arithmetic average-pooling degrades the performance in CNNs by losing crucial information in strong activation values. Also, max-pooling has a problem by ignoring all information except the largest value.  

In addition to max and arithmetic average, there are many other average methods including geometric and harmonic average. Then, how can we find the best average method? 
In training CNN models, it is not practically possible to find a proper average method for different network architectures. This might be the fundamental reason why diverse pooling types are not applied in practice. In order to avoid such limitations, it is desirable to find an optimal average method for pooling layers automatically from training data. 
There have been several variants to the pooling methods, including stochastic pooling rather than deterministic pooling \cite{stocastic_pooling}, and trainable pooling types \cite{lppooling,lee2016generalizing,yu2014mixed}. Even the trainable pooling methods do not cover the average-pooling and the max-pooling as a special case of a generalized method. 

On the other hand, as a general data integration framework, $\alpha$-integration was proposed \cite{Amari2007nc}. $\alpha$-integration integrates positive values and the characteristics of integration is determined by the parameter $\alpha$. It finds out the optimal integration of the input values in the sense of minimizing the $\alpha$-divergence between the integration and the input values. Many average models such as the mixture (or product) of experts model \cite{Jacobs1991nc,Hinton2002nc} can be considered as special cases of $\alpha$-integration \cite{Amari2007nc}. In addition, a training algorithm was proposed to find the best $\alpha$ value from training data for a given tasks \cite{HChoi2010icasspb,HChoi2013nc}. 

In this paper, we propose a new pooling algorithm, {\it $\alpha$-integration pooling} ($\alpha$I-pooling), which applies $\alpha$-integration to the pooling layers in CNNs. $\alpha$I-pooling finds the optimal $\alpha$ values for the pooling layers automatically from training data by back-propagation. So, when we need sub-sampling, we do not have to predefine a specific pooling type. With $\alpha$I-pooling, the model finds the optimal pooling from training data for the task. 

In experiments, $\alpha$I-pooling improves significantly the accuracy of image recognition. In other words, max-pooling might not be always the best pooling method. After training models, we found that the layers have different $\alpha$ values, which means optimal average types for layers could be different. 

The rest of the paper is organized as follows. In section 2, we briefly review $\alpha$-integration and the pooling methods in CNNs. In section 3, we propose $\alpha$I-pooling. In section 4, experiment results confirm our method, followed by conclusion in section 5. 

\section{Background}
In this section, we briefly review $\alpha$-integration \cite{Amari2007nc} and several pooling methods in CNNs.

\subsection{$\alpha$-Integration}
\begin{figure}
\centering
\centerline{\includegraphics[width=3.5in]{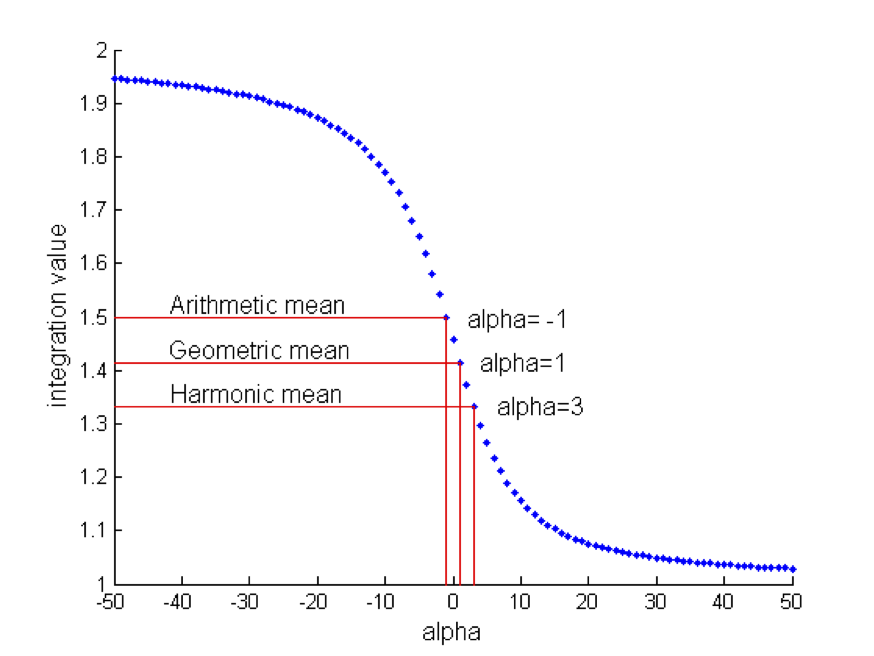}}
\caption{$\alpha$-integration of two values, 1 and 2. As $\alpha$ increases or decreases, the $\alpha$-integration value decreases or increases monotonically and converges 1 or 2, respectively \cite{HChoi2013nc}.}
\label{fig:alpha_mean}
\end{figure}

Given two positive measures of random variable $x$, $m_{1}(x) > 0$ and $m_{2}(x) > 0$, $\alpha$-means is defined by 
\begin{equation} 
\label{eq:alphameans_1}
\widetilde{m}_{\alpha}(x) = f_{\alpha}^{-1} \left( \frac{1}{2} \bigl\{ f_{\alpha}(m_1(x))
+ f_{\alpha}(m_2(x)) \bigr\} \right),
\end{equation}
where $f_{\alpha}(\cdot)$ is a differential monotone function given by
\begin{equation} 
\label{eq:alphameans_2}
f_{\alpha}(z) = \left\{ \begin{array}{ll} z^{\frac{1-\alpha}{2}}, 
& \alpha \neq 1,\\
\log z, & \alpha = 1.
\end{array}\right.
\end{equation}
$\alpha$-mean includes many means as a special case. That is, arithmetic mean, geometric mean, harmonic mean, minimum, and maximum are specific cases of $\alpha$-mean with $\alpha = -1, 1, 3, \infty$ or $-\infty$, respectively, as shown in Fig. \ref{fig:alpha_mean}. As $\alpha$ decreases, the $\alpha$-mean approaches to the larger of $m_{1}(x)$ or $m_{2}(x)$, and when $\alpha=-\infty$, $\alpha$-mean behaves as the max operation. 

Given $M$ positive values $m_1(x),\ldots,m_M(x)$, $\alpha$-mean can be generalized to $\alpha$-integration, which is defined by
\begin{equation} 
\label{eq:alpha_integration}
\tilde{m}_{\alpha}(x) = f_{\alpha}^{-1}\left( \frac{1}{M} \sum_{i=1}^{M} f_{\alpha}(m_{i}(x)) \right) ,
\end{equation}
where we assume that the values have the same weights. 

In most existing works on $\alpha$-integration \cite{Amari2007nc,HChoi2010icasspa,HChoi2010pricai}, the value of $\alpha$ is given in advance rather than learned. To find the optimal $\alpha$ value automatically based on training data, a gradient descent algorithm was proposed for a given task \cite{HChoi2010icasspb,HChoi2013nc}.

\subsection{Max-pooling and Average-pooling}
CNNs are composed of convolutional layers, nonlinear function and pooling layers. Convolutional layers extract patterns from local regions of the input images \cite{book_deeplearning}. Filters in convolution layers have high values when each portion of input image matches to the feature. Then, nonlinear function $f(\cdot)$ is applied to the values, which is ReLU for the most cases. The output of the nonlinear function moves to the pooling layer. Pooling provides positional invariance to the feature, which becomes less sensitive to the precise locations of structures within the image than the original feature maps. This is a crucial transformation for classification tasks.

Despite pooling is an important component for CNNs, max-pooling (or sometimes arithmetic average-pooling) is selected for the most cases without much consideration. Max-pooling chooses only the highest value in the pooling window. Arithmetic average pooing chooses the arithmetic average of all values in the window area. However, it is not guaranteed that the two pooling methods are perfect for all the times. 

In general, the arithmetic average-pooling degrades the performance in CNNs when ReLU is used. Averaging ReLU activation values reduces the high values which might be crucial information, because many zero elements are included in the average. The use of $\tanh$ makes the problems worse by averaging out strong positive and negative activation values with mutual cancellation of each other. Although max-pooling does not suffer from this problem, it has another problem. Max-pooling gives the weight 1 only to the largest value, but gives weights 0 to the other values. That is, it ignores all information except the largest value. Thus, there have been many attempts to find better pooling methods \cite{lppooling,lee2016generalizing,yu2014mixed,stocastic_pooling,B1,B2}.

\subsection{Trainable Pooling}
While most generalized pooling methods are not trainable and defined in advance, there are trainable pooling methods including $\alpha$-pooling \cite{simon2017generalized} and $lp$-pooling \cite{lppooling}. These pooling methods are trainable by back-propagation. 

As in \cite{simon2017generalized}, $\alpha$-pooling is defined by 
\begin{equation}
\alpha\text{-pool}_{\alpha}(\{ x_{i} \}_{i=1}^{N}) = \text{vec} \left( \frac{1}{N} \sum_{i=1}^{N} \alpha\text{-prod} (x_{i}, \alpha) \right),
\end{equation}
and
\begin{equation}
\alpha\text{-prod}(x_i,\alpha) = (\text{sign}(x_i) \circ (|x_{i}| + \epsilon )^{\alpha -1}) x_i^T,
\end{equation}
where $\alpha$ is a trainable parameter to define pooling type. They add small constant value $\epsilon$ to $|x_{i}|$ because $|x_{i}|$ should not be zero. This pooling method has arithmetic average-pooling and bilinear pooling \cite{tenenbaum2000separating} as a special case, but max-pooling is not included. Note that max-pooling is the most often used in CNNs. More importantly, their method was applied only to fully connected layers. In contrast, we propose $\alpha$I-pooling as a more generalized formula to CNNs. 

$Lp$-pooling \cite{lppooling} generalizes pooling operation as $lp$-norm, which is defined by
\begin{equation}
lp\text{-pool}_p(\{x\}_{i=1}^N) = \left( \frac{1}{N} \sum_{i=1}^{N} |x_i|^{p} \right)^{\frac{1}{p}},
\end{equation}
This pooling method has arithmetic average-pooling and max-pooling as a special case. They reparameterize $p$ by 1 + $\log(1+e^{p})$ to satisfy $0<p<1$.

\section{$\alpha$-Integration Pooling}
\label{sec:alpha-pooling}
To find optimal average methods for pooling layers automatically from training data, we propose a generalized pooling method, {\it $\alpha$-integration pooling} ($\alpha$I-pooling),
which applies $\alpha$-integration to the pooling layers. Based on solid mathematical background, our proposed $\alpha$I-pooling is defined by
\begin{equation} 
\label{eq:alpha_pool}
\alpha\text{I-pool}_{\alpha}(\{x_{i} \}_{i=1}^{N}) = f_{\alpha}^{-1}\left( \frac{1}{N} \sum_{i}^{N} f_{\alpha}(x_{i}) \right),
\end{equation}
where $f_{\alpha}$ is the same as Eq. (\ref{eq:alphameans_2}). For kernels with a size of $K_1\times K_2$, $N$ is $K_1 K_2$. We treat the $\alpha$I-pooling's $\alpha$ value as a parameter like other parameters (i.e., weights or bias of the network model), so that $\alpha$ can be trained by back-propagation, although we can train $\alpha$ in different ways from other network parameters. 

For the backpropagation to learn $\alpha$, we can take derivative of objective function $\mathcal{L}$ (cross-entropy) with respect to $\alpha$ for each layer, although it can be done by automatic differentiation as in Pytorch. Suppose that at a layer, there are $Q$ output nodes, $m_1, m_2, \cdots, m_Q$, each of which has $N$ inputs like $x_{q1}, x_{q2}, \cdots, x_{qN}$, then the derivative can be obtained by 
\begin{equation} 
\label{eq:dJ}
\frac{\partial \mathcal{L}}{\partial \alpha} = \sum_{i=1}^Q \frac{\partial\mathcal{L}}{\partial m_i} \frac{\partial m_i}{\partial \alpha} = \sum_{i=1}^Q \delta_i \frac{\partial m_i}{\partial\alpha},
\end{equation}
where $\delta_i$ is obtained from the conventional backpropgation, and 
\begin{equation} 
\frac{\partial m_i}{\partial\alpha} = 
\frac{2m_i}{1-\alpha}
\left\{\frac{\log(\sum_{j=1}^N f_{\alpha}(x_{ij}))}{1-\alpha} 
+\frac{\sum_{j=1}^N \frac{\partial f_{\alpha}(x_{ij})}{\partial \alpha}}{\sum_{j=1}^N f_{\alpha}(x_{ij})} \right\},
\end{equation}
and 
\begin{equation} 
\frac{\partial f_{\alpha}(u)}{\partial \alpha}
   =  -\frac{1}{2}\log(u)u^\frac{1-\alpha}{2}.
\end{equation}
We use gradient descent to update $\alpha$ as
\begin{equation} 
\label{eq:update_a}
\Delta\alpha = - \eta_{\alpha} \frac{\partial \mathcal{L}}{\partial \alpha},
\end{equation}
where $\eta_{\alpha}$ is a learning rate for $\alpha$. By updating $\alpha$ repeatedly with other parameters in the neural network until they converge, we can find the optimal value for $\alpha$.

For using $\alpha$-integration to the pooling layers, we should meet one  constraint: all input values to $\alpha$I-pooling must be positive. This constraint is not a big problem because CNNs use ReLU as an activation function for the most cases. There are no negative values in the output of ReLU. Now, we just need to be careful to avoid zeros, because it is impossible to calculate $\alpha$-integration when zero is included. Therefore, we slightly revise the ReLU function by adding $\epsilon$ to the output of ReLU, which leads to a new activation function, $ReLU^+$ as follows.   
\begin{equation}
    ReLU^+(x) = max(\epsilon, x), 
\end{equation}
where $\epsilon$ is a small positive number, which is set to $10^{-8}$ in our experiments. 

\begin{figure}
\centering
\centerline{\includegraphics[width=10cm]{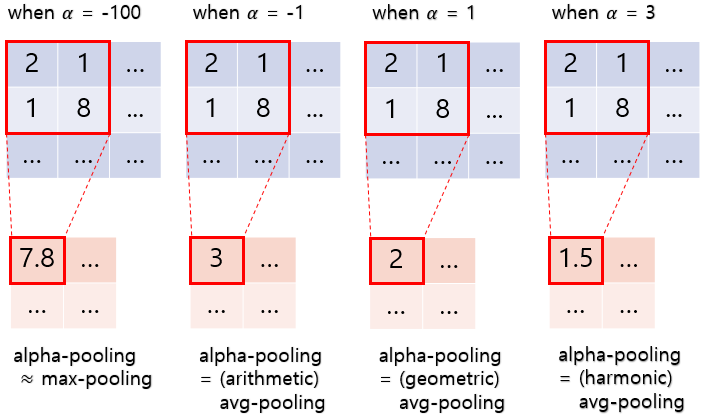}}
\caption{Example of $\alpha$I-pooling results of four positive values: 2, 1, 1, and 8, for a 2$\times$2 pooling window. The outputs of the pooling are different depending on the $\alpha$ value. }
\label{fig:ex_output}
\end{figure}

After applying $ReLU^+$, we $\alpha$-integrate the activation values with current $\alpha$, assuming all the values are positive. Fig. \ref{fig:ex_output} shows an example of how $\alpha$I-pooling works. With different $\alpha$ values, the output of the pooling layer is different. Note that when $\alpha=-\infty$, the integration works as the max operation. 

Now, with $\alpha$I-pooling, a model can find optimal pooling methods from training data for a given task. All pooling layers can share a single $\alpha$ value like a max-pooling, or each layer can have a different pooling type with a different $\alpha$ value. In our experiments, we have different $\alpha$ values for different $\alpha$I-pooling layers.

\section{Experiments}
\label{sec:experiment}
To compare several pooling methods, we present experiment results on four datasets with three CNN models. 

\subsection{Data}
Our experiments are conducted on four datasets: MNIST \cite{mnist_lecun}, CIFAR10, CIFAR100 \cite{cifar-dataset}, and SVHN \cite{svhn-dataset}. 
MNIST includes hand written digit images ($28 \times 28$) of 10 classes (0-9 digits). MNIST splits into two sets: training(60K images) and testing(10K images). CIFAR10 includes images of 10 classes, which has 50K training images and 10K testing images. CIFAR100 is just like CIFAR10 except it has 100 classes. SVHN  includes house number images of 10 classes, which has 73,257 training images and 26,032 testing images.

For CIFAR10, CIFAR100, and SVHN, the input image is $32\times 32$ randomly cropped from a zero-padded $40\times 40$ image. No other data augmentation is applied.

\subsection{Model}
\begin{figure}
\centering
\centerline{
\includegraphics[width=2.5cm]{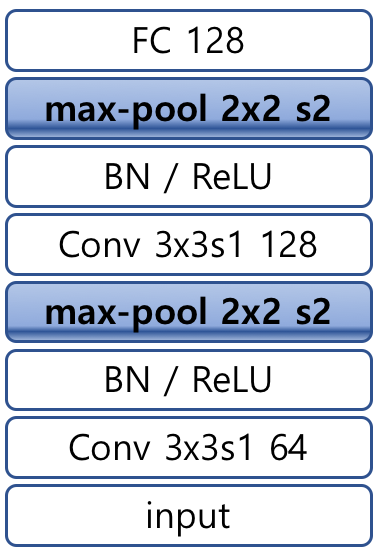}
\hspace{0.4in}
\includegraphics[width=3.1cm]{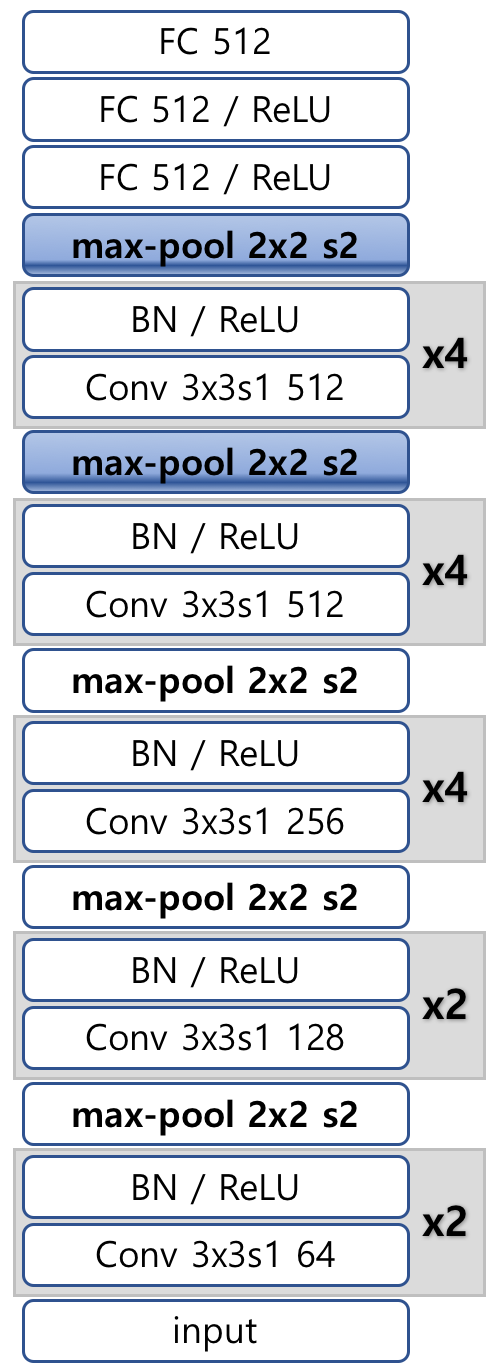}
\hspace{0.4in}
\includegraphics[width=3.9cm]{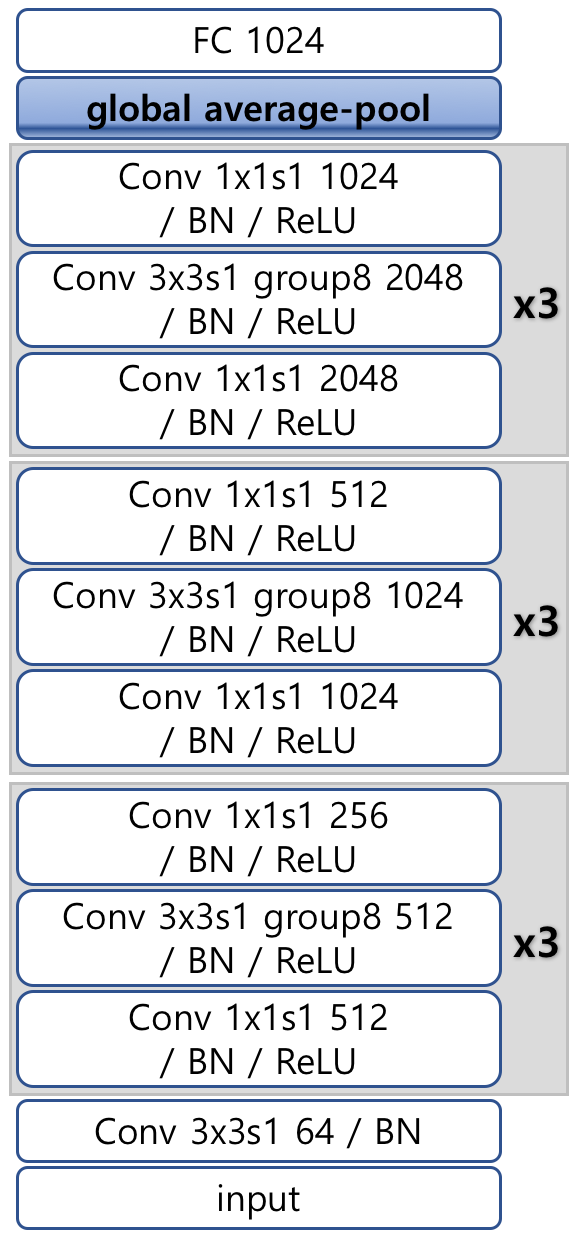}
}
\centerline{(a) \hspace{1.2in} (b) \hspace{1.5in} (c)\hspace{0.35in}}
\caption{Three CNN models: (a) Simple CNN, (b) revised VGG with 19 layers, and (c) ResNeXt with 29 layers, 8 groups, and the width of 64. We apply trainable pooling methods by replacing the two max-pooling layers (or one average-pooling layer for ResNeXt) in the dark boxes with trainable pooling layers, respectively.}
\label{fig:models}
\end{figure}

As shown in Fig. \ref{fig:models}, we take three CNN models for experiments. 
First, we set a simple CNN model to minimize impact of other techniques and to confirm the impact of $\alpha$I-pooling on image recognition. This model consists of two convolutional layers and two pooling layers. Second, we take the VGG model \cite{vgg} to check whether $\alpha$I-pooling works well in complex models. Our experiments are based on a revised VGG model with 19 layers which has 5 max-pooling layers. We replace 2 max-pooling layers with $\alpha$I-pooling layers. 
Third, we take the ResNeXt model \cite{resnext} to know that $\alpha$I-pooling tunes well other type of pooling, since the ResNeXt model includes average-pooling. The model has 29 layers, 8 groups, and the width of 64 of bottleneck. 

To compare trainable pooling methods ($\alpha$-pooling \cite{simon2017generalized}, $lp$-pooling \cite{lppooling}, and our $\alpha$I-pooling), we replace the max-pooling layers or the global average-pooling layer in the dark boxes in Fig. \ref{fig:models} with trainable pooling methods, respectively. 


\subsection{Training}
All models were trained on single GPU with the mini-batch size of 32, the weight decay of 0.0005, and the momentum of 0.9. 

Simple CNN was trained on MNIST, CIFAR10, CIFAR100, and SVHN. We trained the model for 300 epochs with a learning rate of 0.001, and reduced the rate by 0.1 at the 50th and 100th epoch. 
VGG was trained on CIFAR10, CIFAR100, and SVHN. We trained the model for 600 epochs with a learning rate of 0.001. Then, we increased the rate by 10 times at the 10th epoch, and reduced it by 0.1 at the 50th, 100th, 150th, and 300th epochs.
ResNeXt \ref{fig:models} was trained on CIFAR10, CIFAR100, and SVHN. We trained the model with a learning rate of 0.001 for 300 epochs, and increased the rate by 10 times at the 10th epoch, and reduced it by 0.1 at the 50th, 100th, and 150th epoch.


\subsection{Results}

Table \ref{tab:results} presents the experiment results on image recognition tasks with 4 datasets and 3 different CNN models for max-pooling and 3 trainable pooling layers. It confirms that $\alpha$I-pooling works well with different cases. 

\begin{table}
\begin{center}
\caption{Performance in accuracy (\%) of the pooling methods on MNIST, CIFAR10, CIFAR100, and SVHN with three different CNN models. The highest accuracy for each case is marked in bold, and the ones which are worse than max-pooling in gray.}
\label{tab:results}
\begin{tabular}{ |l|l||c|c|c|c| }
\hline
\multicolumn{2}{|c||}{Experiment case}&
\multicolumn{4}{c|}{Pooling method} \\
\hline
Dataset & Model & max-pool & $\alpha$-pool \cite{simon2017generalized} & $Lp$-pool \cite{lppooling} & ~$\alpha$I-pool~ \\
\hline
\hline
\small{MNIST}    & \small{SimpleCNN~}& 98.28 & 98.61 & 98.44 & \textbf{98.62}   \\
\hline
\small{CIFAR10}  & \small{SimpleCNN} & 72.52 & \textbf{83.74} & 82.84 & 83.57  \\
\small{CIFAR10}  & \small{VGG}       & 92.38 & \textcolor{gray}{91.18} & 93.15 & \textbf{93.73}    \\
\small{CIFAR10}  & \small{ResNeXt}   & 95.93 & 96.00 & \textbf{96.18} & \textbf{96.18}   \\
\hline
\small{CIFAR100~}& \small{SimpleCNN} & 58.20 & \textcolor{gray}{57.68} & 58.43 & \textbf{58.80}   \\
\small{CIFAR100} & \small{VGG}       & 72.47 & \textcolor{gray}{71.76} & \textcolor{gray}{71.52} & \textbf{72.51} \\
\small{CIFAR100} & \small{ResNeXt}   & 77.75 & 79.58 & \textcolor{gray}{66.59} & \textbf{79.74} \\
\hline
\small{SVHN}     & \small{SimpleCNN} & 88.00 & \textcolor{gray}{87.98} & 88.16 & \textbf{89.02}   \\
\small{SVHN}     & \small{VGG}       & 95.80 & 96.20 & 96.29 & \textbf{96.87} \\
\small{SVHN}     & \small{ResNeXt}   & 96.87 & \textcolor{gray}{96.67} & \textbf{97.13}    &   97.12   \\
\hline
\end{tabular}
\end{center}
\end{table}

First of all, $\alpha$I-pooling outperforms max-pooling for all cases, which implies that our method finds a better pooling method for each layer than max-pooling for all layers. If max-pooling were optimal for the tasks, the variable $\alpha$ of $\alpha$I-pooling would converge to $-\infty$. In Table \ref{tab:alpha_values}, we can see that all $\alpha$ variables converge to certain values between -6 and 1.

Moreover, $\alpha$I-pooling shows better performance than other trainable pooling methods (i.e., $\alpha$-pooling and $lp$-pooling) in most cases. Even when $\alpha$I-pooling is not the best, the accuracy is close to the best one. Note that the other two trainable pooling methods are worse than max-pooling in 6 cases: 5 cases for $\alpha$-pooling and 2 cases for $Lp$-pooling, which are marked in gray in Table \ref{tab:results}. The consistent improvement by $\alpha$I-pooling comes from solid mathematical support of $\alpha$-integration, so that the pooling type comes close to an optimal pooling type for individual tasks. 


\begin{table}
\begin{center}
\caption{Trained $\alpha$ values for $\alpha$I-pooling on 4 datasets with 3 CNN models. $\alpha_1$ is for the $\alpha$I-pooling layer closer to input than $\alpha_2$ in Fig. \ref{fig:models}. Note that ResNeXt has only one $\alpha$I-pooling layer.}
\label{tab:alpha_values}
\begin{tabular}{ |l|l|| c | c | }
\hline
\multicolumn{2}{|c||}{Experiment case}&
\multicolumn{2}{c|}{Trained $\alpha$ values} \\
\hline
Dataset & Model & ~$\alpha_1$~ & ~$\alpha_2$~ \\
\hline
\hline
MNIST    & SimpleCNN~& ~~~ -0.8987 ~~~ & ~~~ -2.6298 ~~~   \\
\hline
CIFAR10  & SimpleCNN & -1.5639 & -2.2038  \\
CIFAR10  & VGG       &  0.3342 & -2.6644    \\
CIFAR10  & ResNeXt   &  0.0047 & N/A  \\
\hline
CIFAR100~& SimpleCNN & -1.6078 & -2.9101   \\
CIFAR100 & VGG       & -2.0166 & -0.3336 \\
CIFAR100 & ResNeXt   &  0.0068 & N/A \\
\hline
SVHN     & SimpleCNN & -1.1101 & -1.2543  \\
SVHN     & VGG       & -1.2616 & -2.1639 \\
SVHN     & ResNeXt   & -5.6052 & N/A  \\
\hline
\end{tabular}
\end{center}
\end{table}

Table \ref{tab:alpha_values} shows that $\alpha$ for different layers converge to different values. This implies that each layer has a different optimal pooling type, because its role is different depending on datasets and models. Also, it implies that there is no single optimal pooling type.

Interestingly, the $\alpha$ value for the SVHN-ResNeXt case is -5.6052 which is not the average-pooling as the original ResNeXt model proposes, while for CIFAR10-ResNeXt or CIFAR100-ResNeXt, the $\alpha$ values indicate that the pooling is close to average-pooling as the original ResNeXt model proposes. Note that SVHN includes simple number images of 10 classes, while CIFAR is regular image dataset. From the $\alpha$ values for ResNeXt, we can see that the optimal pooling type by $\alpha$I-pooling depends on the dataset.

\section{Conclusion}
\label{sec:conclusion}
In this paper, we questioned about the pooling methods, which find a sub-sampled value within a window. We proposed $\alpha$I-pooling to include the previous pooling methods as a special case. The parameter $\alpha$ of $\alpha$I-pooling is trainable from training data by back-propagation, and the converged $\alpha$ value determines the pooling type automatically.

Experiment results confirm that $\alpha$I-pooling improves performance compared to max-pooling and other trainable pooling methods. Also, each pooling layer has different $\alpha$ value, suggesting that there is no single optimal pooling type for all cases. As future works, we can analyze the meaning of different $\alpha$ values in detail, and investigate effects of $\alpha$I-pooling when applied with other learning techniques.

\section{Acknowledgement}
\label{sec:acknowledgement}
This work was supported by Institute for Information \& communications Technology Promotion(IITP) grant funded by the Korea government(MSIT) (No.2018-0-00749, Development of virtual network management technology based on artificial intelligence) and Basic Science Research Program through the National Research Foundation of Korea (NRF) funded by the Ministry of Education (2017R1D1A1B03033341).

%
%
\bibliographystyle{jponew}

\bibliography{refs}

\begin{thebibliography}{10}
\providecommand{\url}[1]{\texttt{#1}}
\providecommand{\urlprefix}{URL }
\providecommand{\doi}[1]{https://doi.org/#1}

\bibitem{Amari2007nc}
Amari, S.: Integration of stochastic models by minimizing $\alpha$-divergence.
  Neural Computation  \textbf{19},  2780--2796 (2007)

\bibitem{HChoi2013nc}
Choi, H., Choi, S., Choe, Y.: Parameter learning for alpha-integration. Neural
  Computation  \textbf{25}(6),  1585--1604 (2013)

\bibitem{HChoi2010icasspb}
Choi, H., Choi, S., Katake, A., Choe, Y.: Learning alpha-integration with
  partially-labeled data. pp. 2058--2061. Dallas, TX (2010)

\bibitem{HChoi2010icasspa}
Choi, H., Katake, A., Choi, S., Choe, Y.: Alpha-integration of multiple
  evidence. In: International Conference on Acoustics, Speech, and Signal
  Processing (ICASSP). pp. 2210--2213. Dallas, TX (2010)

\bibitem{B2}
Douze, M., Revaud, J., Schmid, C., Jegou, H.: Stable hyper-pooling and query
  expansion for event detection. In: International Conference on Computer
  Vision (ICCV). pp. 1825--1832 (2013). \doi{10.1109/ICCV.2013.229},
  \url{https://doi.org/10.1109/ICCV.2013.229}

\bibitem{B1}
Gao, Y., Beijbom, O., Zhang, N., Darrell, T.: Compact bilinear pooling. In:
  Conference on Computer Vision and Pattern Recognition (CVPR). pp. 317--326
  (2016). \doi{10.1109/CVPR.2016.41},
  \url{https://doi.org/10.1109/CVPR.2016.41}

\bibitem{book_deeplearning}
Goodfellow, I., Bengio, Y., Courville, A.: Deep Learning. MIT Press (2016),
  \url{http://www.deeplearningbook.org}

\bibitem{lppooling}
G{\"{u}}l{\c{c}}ehre, {\c{C}}., Cho, K., Pascanu, R., Bengio, Y.: Learned-norm
  pooling for deep feedforward and recurrent neural networks. In: European
  Conference on Machine Learning (ECML). pp. 530--546 (2014).
  \doi{10.1007/978-3-662-44848-9\_34},
  \url{https://doi.org/10.1007/978-3-662-44848-9\_34}

\bibitem{Hinton2002nc}
Hinton, G.E.: Training products of experts by minimizing contrastive
  divergence. Neural Computation  \textbf{14},  1771--1800 (2002)

\bibitem{hubel1962receptive}
Hubel, D.H., Wiesel, T.N.: Receptive fields, binocular interaction and
  functional architecture in the cat's visual cortex. The Journal of physiology
   \textbf{160}(1),  106--154 (1962)

\bibitem{Jacobs1991nc}
Jacobs, R.A., Jordan, M.I., Nowlan, S.J., Hinton, G.E.: Adaptive mixtures of
  local experts. Neural Computation  \textbf{3},  79--81 (1991)

\bibitem{cifar-dataset}
Krizhevsky, A., Hinton", G.: Learning multiple layers of features from tiny
  images. Technical Report  (2009)

\bibitem{alexnet}
Krizhevsky, A., Sutskever, I., Hinton, G.E.: Imagenet classification with deep
  convolutional neural networks. In: Advances in Neural Information Processing
  Systems (NIPS). pp. 1106--1114 (2012),
  \url{http://papers.nips.cc/paper/4824-imagenet-classification-with-deep-convolutional-neural-networks}

\bibitem{mnist_lecun}
LeCun, Y., Bottou, L., Bengio, Y., Haffner, P.: Gradient-based learning applied
  to document recognition. Proceedings of the {IEEE}  \textbf{86}(11),
  2278--2324 (1998)

\bibitem{lee2016generalizing}
Lee, C.Y., Gallagher, P.W., Tu, Z.: Generalizing pooling functions in
  convolutional neural networks: Mixed, gated, and tree. In: Artificial
  intelligence and statistics. pp. 464--472 (2016)

\bibitem{svhn-dataset}
Netzer, Y., Wang, T., Coates, A., Bissacco, A., Wu, B., Ng, A.Y.: Reading
  digits in natural images with unsupervised feature learning  (2011)

\bibitem{simon2017generalized}
Simon, M., Gao, Y., Darrell, T., Denzler, J., Rodner, E.: Generalized orderless
  pooling performs implicit salient matching. In: Proceedings of the IEEE
  international conference on computer vision. pp. 4960--4969 (2017)

\bibitem{vgg}
Simonyan, K., Zisserman, A.: Very deep convolutional networks for large-scale
  image recognition. CoRR  \textbf{abs/1409.1556} (2014),
  \url{http://arxiv.org/abs/1409.1556}

\bibitem{inception}
Szegedy, C., Ioffe, S., Vanhoucke, V., Alemi, A.A.: Inception-v4,
  inception-resnet and the impact of residual connections on learning. In:
  Association for the Advancement of Artificial Intelligence (AAAI). pp.
  4278--4284 (2017),
  \url{http://aaai.org/ocs/index.php/AAAI/AAAI17/paper/view/14806}

\bibitem{tenenbaum2000separating}
Tenenbaum, J.B., Freeman, W.T.: Separating style and content with bilinear
  models. Neural computation  \textbf{12}(6),  1247--1283 (2000)

\bibitem{resnext}
Xie, S., Girshick, R., Doll{\'a}r, P., Tu, Z., He, K.: Aggregated residual
  transformations for deep neural networks. In: Proceedings of the IEEE
  conference on computer vision and pattern recognition. pp. 1492--1500 (2017)

\bibitem{yu2014mixed}
Yu, D., Wang, H., Chen, P., Wei, Z.: Mixed pooling for convolutional neural
  networks. In: International Conference on Rough Sets and Knowledge
  Technology. pp. 364--375. Springer (2014)

\bibitem{stocastic_pooling}
Zeiler, M.D., Fergus, R.: Stochastic pooling for regularization of deep
  convolutional neural networks. CoRR  \textbf{abs/1301.3557} (2013),
  \url{http://arxiv.org/abs/1301.3557}

\end{thebibliography}
\end{document}